\documentclass{article}

 \usepackage[preprint]{neurips_2026}

\usepackage[utf8]{inputenc} 
\usepackage[T1]{fontenc}    
\usepackage{hyperref}       
\usepackage{url}            
\usepackage{booktabs}       
\usepackage{amsfonts}       
\usepackage{nicefrac}       
\usepackage{microtype}      
\usepackage{xcolor}         
\usepackage{xspace}
\usepackage{makecell}

\usepackage{pifont}
\usepackage{graphicx}
\newcommand{\tempglitch}{\emph{TempGlitch}\xspace}
\usepackage[most]{tcolorbox}
\newtcolorbox{rqbox}{
  colback=black!2,
  colframe=black!30,
  boxrule=0.4pt,
  arc=3pt,              
  left=4pt,right=4pt,top=3pt,bottom=3pt,
  boxsep=2pt
}
\usepackage{float} 
\usepackage{array}

\usepackage[most]{tcolorbox}
\newcommand{\promptsection}[2]{%
\begin{tcolorbox}[
    colback=#1!5,
    colframe=#1!65!black,
    boxrule=0.6pt,
    arc=2pt,
    left=6pt,
    right=6pt,
    top=5pt,
    bottom=5pt,
    title=\texttt{#2},
    fonttitle=\bfseries\ttfamily\small,
    coltitle=white,
    colbacktitle=#1!75!black
]
}

\newcommand{\cmark}{\textcolor{green!50!black}{\ding{51}}}
\newcommand{\xmark}{\textcolor{red}{\ding{55}}}

\title{\tempglitch: Evaluating Vision-Language Models for Temporal Glitch Detection in Gameplay Videos}

\author{%
  Yakun Yu$^{1}$ \quad Ashley Wiens$^{1}$ \quad Adrián Barahona-Ríos$^{2}$ \quad Benedict Wilkins$^{2}$\\ \textbf{Saman Zadtootaghaj$^{2}$ \quad Nabajeet Barman$^{2}$ \quad Cor-Paul Bezemer$^{1}$}\\
  \texttt{ 
  \{yakun,bezemer\}@ualberta.ca,} \\
  \texttt{ \{Saman.Zadtootaghaj,Nabajeet.Barman\}@sony.com} \\
  {}$^1$ University of Alberta \quad {}$^2$ Sony Interactive Entertainment
}

\begin{document}

\maketitle

\begin{abstract}
Vision-language models (VLMs) are increasingly being explored for video game quality assurance, especially gameplay glitch detection. 
Most existing evaluations, however, treat glitches as static visual anomalies, asking models to detect failures from a single frame. 
We argue that this framing misses a key distinction: some glitches are \emph{spatial} and visible in an isolated frame, whereas others are \emph{temporal} and become evident only through changes across ordered frames. 
A preliminary study confirms this gap, showing that temporal glitches are substantially harder for VLMs to detect than spatial ones. 
To enable systematic evaluation of this underexplored setting, we introduce \tempglitch, a controlled gameplay video benchmark for temporal glitch detection. 
\tempglitch covers five temporal glitch types with balanced per-category samples, together with paired glitch-free videos that enable reliable binary evaluation.
We evaluate 12 proprietary and open-weight VLMs across multiple frame-sampling settings. 
Our results show that current VLMs remain near chance on \tempglitch, often collapsing into either overly conservative behavior that misses most glitches or overly sensitive behavior that flags clean videos as glitchy. 
Moreover, denser frame sampling and larger model size do not reliably resolve these failures. 
\tempglitch provides a focused testbed for temporal reasoning, robust gameplay understanding, and automated glitch detection with VLMs.
Code and data are available at \href{https://asgaardlab.github.io/TempGlitch/}{the project website}.
    
\end{abstract}

\section{Introduction}
\label{sec:intro}
The video game industry has expanded rapidly over the past decade and now represents one of the largest and most commercially significant sectors of interactive entertainment~\cite{taesirivideogameqa,yu2026resp}.
At the same time, modern game development has become increasingly complex, requiring the coordinated design of large-scale visual assets, physics systems, interaction logic, and platform-specific optimizations~\cite{politowski2021game,lear2019asset,chueca2024consolidation}.
As production pipelines grow in scale and sophistication, maintaining quality throughout development becomes correspondingly more challenging.

\begin{table}[t]
\centering
\small
\setlength{\tabcolsep}{5pt}
\caption{GPT-5 glitch detection performance under different frame sampling settings.
``Glitchy'' is treated as the positive class. Spatial and temporal recall are computed on 50 spatial-glitch and 50 temporal-glitch videos from PhysGame~\cite{cao2024physgame}, respectively. Specificity is computed on 100 glitch-free videos from VideoGameQA-Bench~\cite{taesirivideogameqa}. Higher values indicate better performance.}
\label{tab:gpt5_binary_bug_detection}
\begin{tabular}{lccccccc}
\toprule
& \multicolumn{3}{c}{\textbf{Glitchy videos}} & \textbf{Glitch-free videos} & \multicolumn{3}{c}{\textbf{Overall}} \\
\cmidrule(lr){2-4} \cmidrule(lr){5-5} \cmidrule(lr){6-8}
\makecell{Sampling\\strategy}
& \makecell{Spatial\\Recall }
& \makecell{Temporal\\Recall }
& Gap (S$-$T) 
& Specificity 
& Precision 
& F1 
& Accuracy \\
\midrule
I-frames
& 0.72
& 0.48
& +0.24
& 0.89
& 0.84
& 0.70
& 0.74 \\
1 FPS
& 0.74
& 0.68
& +0.06
& 0.81
& 0.79
& 0.75
& 0.76\\
2 FPS
& 0.80
& 0.68
& +0.12
& 0.80
& 0.80
& 0.76
& 0.77 \\
\bottomrule
\end{tabular}
\end{table}

A central challenge in this process is quality assurance (QA), in particular, glitch detection.
Visual glitches can degrade player experience, reduce perceived product quality, and delay deployment if they are discovered late in the development cycle~\cite{backus2025players, lin2019empirical, lu2025automated}.
Traditionally, glitch detection relies heavily on manual review by QA teams, who inspect gameplay footage or interact with builds to identify anomalous behavior.
Although effective in some cases, this process is expensive, time-consuming, and difficult to scale, especially for modern titles that generate large volumes of gameplay data across diverse scenes, devices, and settings~\cite{chang2019reveal, taesiri2022clip, taesiri2024searching, politowski2021survey, politowski2022towards}.

Recent advances in vision-language models (VLMs)~\cite{gan2022vision, li2025survey, dai2023instructblip, wang2025internvl, danish2025comprehensive} have created promising opportunities for automating gameplay glitch detection~\cite{taesirivideogameqa, taesiri2025videogamebunny,cao2024physgame, yu2026resp}.
Their strong multimodal reasoning capabilities suggest that they could assist QA pipelines by interpreting complex visual evidence and identifying abnormal events directly from gameplay data.
However, current performance remains limited~\cite{cao2024physgame, yu2026resp}.
A key issue is that prior work often treats gameplay glitch detection as a single, uniform problem, ignoring that different types of glitches vary substantially in how difficult they are for VLMs to detect. In particular, glitches that unfold over time pose fundamentally different challenges than those identifiable from a single frame.
A second issue is the absence of datasets designed to capture and isolate these more challenging cases, especially those that require temporal reasoning rather than static visual recognition.

\begin{figure}[th!]
  \centering
  \includegraphics[width=\textwidth]{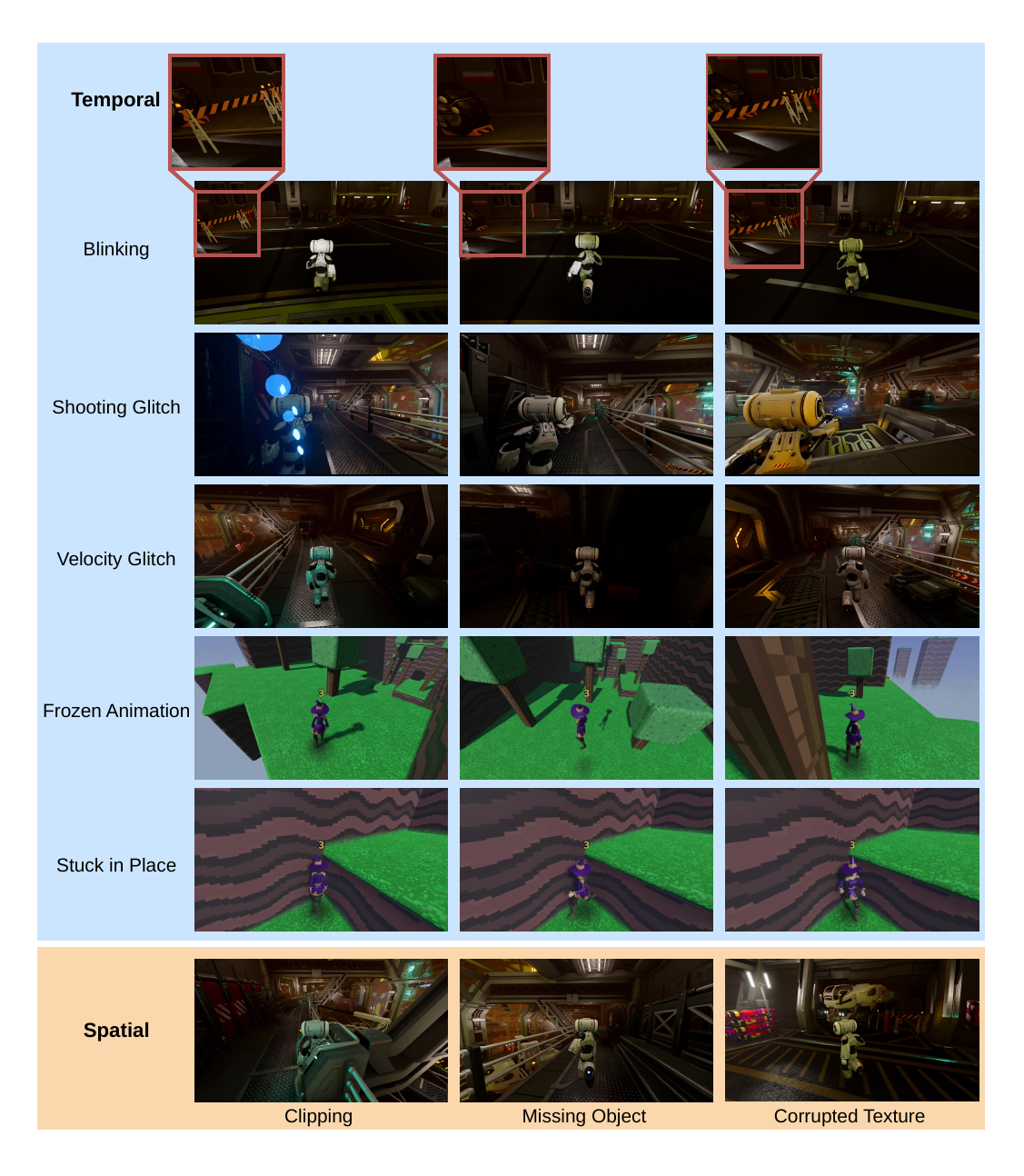}
  \vspace{-30pt}
  \caption{\textbf{Spatial versus temporal glitches.} The top part shows temporal glitches represented by several frames from one video: \emph{Blinking} glitch refers to an object intermittently appearing and disappearing across frames; \emph{Shooting glitch} indicates an artefact when the firing effect occurs at an incorrect position relative to the weapon and target; \emph{Velocity glitch} are implausible changes in character motion velocity, with displacement varying abruptly across adjacent frames; \emph{Frozen animation} indicates that the character is translating through the scene but its pose remains unnaturally fixed; and \emph{Stuck in place} shows the character repeatedly attempting to move while remaining pinned to essentially the same location. The bottom row shows a few examples of spatial glitches~\cite{yu2026resp}, which can be identified from a single frame. Best viewed when zoomed. 
  }
  \label{fig:data}
\end{figure}

In this paper, we argue that there are two major groups of gameplay glitches: \textit{spatial glitches} and \textit{temporal glitches}, as shown in Figure~\ref{fig:data}.
Spatial glitches can be identified from a single frame, because the anomaly is visually apparent in an isolated image.
Examples include clipping, missing objects, or corrupted textures.
Temporal glitches, in contrast, can only be identified from a temporally ordered sequence of frames, because the abnormality lies in how the scene changes over time.
Examples include intermittent disappearance, flickering, or other unstable temporal behaviors.
We hypothesize that temporal glitches are substantially more difficult for current VLMs to detect than spatial glitches.

To test this hypothesis, we perform a preliminary study using GPT-5~\cite{singh2025openai}, one of the strongest VLMs.
Because no existing dataset is explicitly organized around the distinction between spatial and temporal glitches, we derive a curated evaluation subset from the PhysGame~\cite{cao2024physgame} dataset.
We first use GPT-5 to classify the ground-truth bug descriptions of all PhysGame videos as spatial or temporal, together with a confidence score.
We then retain high-confidence samples and manually filter misclassified cases, resulting in a curated subset of 50 spatial and 50 temporal glitch videos.
We further include 100 glitch-free videos from VideoGameQA-Bench~\cite{taesirivideogameqa} to measure if the model is biased toward predicting every video as glitchy.
We evaluate GPT-5 under multiple frame-sampling strategies chosen to ensure that glitchy moments are covered.
Table~\ref{tab:gpt5_binary_bug_detection} shows that GPT-5 is consistently better at detecting spatial glitches than temporal glitches across all sampling strategies.
High specificity on glitch-free videos indicates that this gap reflects a systematic difference rather than random variation.

These observations motivate our focus on temporal glitch detection. Temporal glitches constitute a distinct and systematically harder challenge for current VLMs, yet no benchmark enables their direct study. To fill this gap, we introduce \tempglitch, a benchmark tailored to temporal glitch detection in gameplay videos. \tempglitch enables systematic evaluation of VLMs on glitches that require temporal reasoning, providing a foundation for understanding model failures and informing future improvements.
Our paper makes the following contributions:
\begin{enumerate}
\item \textbf{A spatial--temporal formulation of gameplay glitches.}
    We formalize the distinction between spatial glitches, which can be detected from isolated frames, and temporal glitches, which require reasoning over ordered visual evidence.

    \item \textbf{A dedicated temporal glitch benchmark.}
    We introduce \tempglitch, a controlled gameplay video benchmark of 750 temporal glitch videos across 5 glitch types, paired with 750 glitch-free videos for reliable binary evaluation.

    \item \textbf{An evaluation of 12 VLMs on \tempglitch.}
    We evaluate 12 proprietary and open-weight VLMs on \tempglitch, providing a systematic snapshot of current model capabilities on temporal gameplay glitch detection.

    \item \textbf{New diagnostic findings about VLM failure modes.}
    Our results show that current VLMs remain near chance accuracy on \tempglitch, often collapsing toward either overly conservative or overly sensitive predictions.
    We further find that model scale alone does not reliably solve temporal glitch detection.
\end{enumerate}

\section{Related Work}
Existing resources for gameplay glitch detection can be broadly grouped into image-level and video-level benchmarks.
On the image side, GlitchBench~\citep{taesiri2024glitchbench} introduces 330 glitch-free and 593 glitchy screenshots taken from 205 games to test whether large multimodal models can detect and interpret visual anomalies. 
VideoGameBunny~\cite{taesiri2025videogamebunny} substantially expands this direction by releasing a large-scale dataset with 185,259 images from 413 titles and 389,565 image--instruction pairs, aimed at improving game-image understanding.
These datasets have played an important role in establishing gameplay glitch detection as a multimodal reasoning problem, but they primarily emphasize glitches that can be identified from isolated frames.
Video-level benchmarks are relatively limited.
GameBugDescriptions~\cite{taesiri2022large} formulates glitch detection as question answering over textual descriptions of gameplay events, with 167 glitchy gameplay videos and 334 QA pairs across 8 games. 
VideoGameQA-Bench~\cite{taesirivideogameqa} is a broad QA benchmark covering 2,236 image-based samples and 1,200 video-based
samples from more than 800 games and 9 synthetic game scenes.
PhysGame~\cite{cao2024physgame} focuses specifically on physical commonsense violations in gameplay videos and contains 880 glitchy videos spanning four physical domains and 12 categories. 
Together, these datasets have advanced the study of gameplay video glitch detection, but none is designed specifically to isolate temporal glitches as a first-class benchmark target. Table~\ref{tab:benchmark_comparison} lists the differences between our dataset and the existing datasets for video game glitch detection.

VLMs have recently shown matching or exceeding human performance on various computer vision tasks~\cite{liu2023visual, liu2024mmbench, wang2024cogvlm, lin2024video, 11092508, saxena2026vlm}. 
However, current evidence suggests that VLM performance on gameplay glitch detection remains far from satisfactory, especially in video settings~\cite{cao2024physgame, taesirivideogameqa}. 
VideoGameQA-Bench~\cite{taesirivideogameqa} reports that frontier VLMs perform noticeably worse on video glitch detection than on many simpler game-image tasks. PhysGame~\cite{cao2024physgame} likewise shows that even state-of-the-art VLMs have substantial difficulty on gameplay videos with physical commonsense violations, with open models lagging proprietary ones by a wide margin. 
Our empirical findings suggest one likely reason for this broader weakness: temporal glitches are systematically harder to detect than spatial glitches, yet existing benchmarks do not explicitly isolate this axis of difficulty. As a result, poor performance on video glitch detection may partly reflect the fact that temporally grounded glitches are underrepresented, entangled with other factors, or not analyzed separately.
Therefore, the central contribution of \tempglitch is not only a new dataset, but also a new evaluation perspective: by treating temporal glitches as a distinct benchmark problem, \tempglitch enables systematic analysis of the temporally grounded failures that current VLMs handle poorly.

\begin{table}[t]
\centering
\caption{A summary of existing game glitch detection benchmarks.}
\label{tab:benchmark_comparison}
\resizebox{\linewidth}{!}{
\begin{tabular}{lccrrrrc}
\toprule
\textbf{Benchmarks}
& \textbf{Image-Based}
& \textbf{Video-Based}
& \textbf{\# Videos}  
& \textbf{\# Glitchy videos} 
& \textbf{\# Glitch-free videos} 
& \textbf{Temporal glitch only} 
\\
\midrule
GlitchBench~\cite{taesiri2024glitchbench}     & \cmark   & \xmark &0       & 0 & 0 & \xmark \\
VideoGameBunny~\cite{taesiri2025videogamebunny}  & \cmark    & \xmark & 0     & 0 & 0 & \xmark \\
GameBugDescriptions~\cite{taesiri2022large}        & \xmark    & \cmark    & 167 & 167 & 0 & \xmark \\
VideoGameQA-Bench~\cite{taesirivideogameqa} & \cmark    & \cmark & 1,000    & 500 & 500 & \xmark \\
PhysGame~\cite{cao2024physgame}           & \xmark   & \cmark & 880     & 880 &0 & \xmark  \\
\tempglitch (ours)          & \xmark  & \cmark  & 1,500      & 750 & 750 & \cmark \\
\bottomrule
\end{tabular}
}
\end{table}

\section{The \tempglitch Benchmark}

\subsection{Task Formulation}
We use \emph{glitch} as an umbrella term for unintended, user-visible anomalies that occur during gameplay.
In this work, we focus on glitches that are visually evidenced in gameplay videos.
We group such glitches into two broad families based on how they manifest over time: \emph{spatial glitches} and \emph{temporal glitches}, see Figure~\ref{fig:data}.

\paragraph{Spatial glitches.}
Spatial glitches are anomalies that can be identified from a single still frame of a video.
Examples include, but are not limited to:
\begin{itemize}
    \item \textbf{Clipping}: a character or object visibly intersects or passes through a solid surface, or geometry overlaps in a way that would not occur under correct depth or occlusion.
    \item \textbf{Missing Object}: parts of the character or object are unexpectedly missing.
    \item \textbf{Corrupted texture}: a surface exhibits clearly broken, missing, or incorrect texturing.
\end{itemize}

\paragraph{Temporal glitches.}
Temporal glitches cannot be reliably identified from a single frame,  instead requiring the analysis of multiple temporally ordered frames. 
Examples of glitches that fall into this category include the following:
\begin{itemize}
    \item \textbf{Blinking}: an object intermittently appears and disappears across successive frames.
    Importantly, the scene should remain visually plausible both when the object is visible and when it is absent; otherwise, the glitch may be detectable from a single frame.
    For example, if an entire floor disappears, the glitch is effectively spatial rather than temporal.

    \item \textbf{Shooting glitch}: firing or projectile behavior unfolds incorrectly over time, violating the expected temporal progression of aiming, shooting, and projectile motion.
    Without multiple frames, the same effect could be believably achieved by firing then turning rapidly.

    \item \textbf{Velocity glitch}: an object or character exhibits abnormal motion dynamics over time, such as implausible speed changes, sudden displacement, or inconsistent movement.
    Individual frames should remain visually plausible: the character animation and surrounding environment should appear normal, while the glitch becomes evident only from the motion pattern across frames.

    \item \textbf{Frozen animation}: a character or object remains locked in the same pose or animation frame while the game state continues to evolve.
    The key signal is an animation-state failure: the entity may still translate, rotate, or interact with the scene, but its pose does not update naturally over time.
    Thus, any single frame may look plausible, while the ordered sequence reveals that the animation is unnaturally fixed.

    \item \textbf{Stuck in place}: a character or object fails to change position despite gameplay context indicating that it should move.
    The key signal is a displacement failure: the entity may continue playing a walking or running animation or receive movement input, but its world position remains fixed.
    Thus, the sequence reveals persistent lack of movement, often distinguishable from normal obstruction only by observing the surrounding context and repeated motion attempts.

\end{itemize}

Existing gameplay glitch benchmarks span many games, rendering styles, and camera behaviors, but they are not designed to isolate temporal glitches.
As a result, temporally grounded failure modes are often entangled with cross-environment variation, making systematic analysis difficult.
Moreover, existing datasets typically provide only limited samples for any individual temporal glitch type, which makes category-level evaluation underpowered.
To address this gap, we introduce \tempglitch, a controlled synthetic benchmark specifically designed for temporal glitch detection in gameplay videos.
Its main advantages are: (i) controlled rendering, camera, and scene conditions within a small set of game environments, and (ii) substantial per-category coverage for temporally grounded glitches, enabling reliable benchmarking and fine-grained analysis.

\subsection{Data Collection}
We built \tempglitch using the open-source Godot engine\footnote{\url{https://godotengine.org}}, using the Third Person Shooter (TPS) demo asset\footnote{\url{https://godotengine.org/asset-library/asset/2710}} and the Platformer 3D Demo asset\footnote{\url{https://godotengine.org/asset-library/asset/2748}} as base environments.
For additional visual diversity, we modified the Platformer asset to use an alternative character model\footnote{\url{https://quaternius.com/packs/ultimatemodularwomen.html}}.
We modified the source code to inject temporal glitches through flexible in-engine mechanisms, allowing the same environments, assets, and gameplay setups to exhibit both normal and temporally glitchy behavior.
For each glitch category, we also designed glitch-free behaviors that closely resemble the glitchy cases, so that models should be expected to distinguish temporal failures rather than rely on unrelated scene differences.
We generated data across five temporal glitch categories and matched normal behaviors:

\begin{enumerate}
    \item \textbf{Blinking}: created by toggling the visibility of objects during gameplay.
   We attached a script to selected objects to alternate their visibility at a fixed interval when the glitch is enabled.
    This produces a temporal flickering effect while leaving movement and interaction logic unchanged.
    The corresponding normal videos keep one or two specific objects visible for most of the clip, matching the camera style and object layout of the glitchy videos.

    \item \textbf{Shooting glitch}: based on a glitch already present in the TPS asset, requiring no additional modification.
    The glitch occurs when shooting within a very small temporal window after aiming, causing the bullet trajectory to fire behind the player rather than toward the crosshair.
    The corresponding normal videos also contain shooting events, unlike other categories that focus primarily on movement.
    Some normal clips include quick rotations to capture non-glitchy situations in which a bullet may visually appear behind the player due to legitimate camera or body motion.

    \item \textbf{Velocity glitch}: created by altering the player's movement velocity during otherwise normal navigation.
    We added a modifier that can be toggled to the character movement script that increases the player speed to an extreme value, often causing the traversal to complete within a single frame and visually resemble teleportation.
    The corresponding normal videos contain long straight-line walking segments with similar start and end positions, matching the intended traversal path without abnormal velocity.
    
    \item \textbf{Frozen animation}: implemented by interrupting character animation playback while preserving the rest of the game logic.
    We attached a script that can be toggled to the player animation controller that pauses the current animation state, causing the character model to remain visually frozen even as the underlying gameplay state continues to update.
    The corresponding normal videos contain natural-looking movement and occasional moments of stillness, reflecting legitimate cases where animation may pause briefly.

    \item \textbf{Stuck in place}: implemented by temporarily constraining the player's position despite continued input and animation updates.
    We attached a script to the character that temporarily disables physics processing, preventing displacement while the rest of the gameplay scene remains active.
    This glitch spans both base environments, with glitchy and normal videos split evenly between them.
    The corresponding normal videos include cases where the character repeatedly runs into walls during otherwise standard movement.
    This is particularly useful in the TPS environment, where the running animation continues to play, visually resembling the glitchy behavior except that the normal case contains a visible obstacle causing the lack of movement.
\end{enumerate}

We recorded both glitchy and glitch-free gameplay clips of 5--32 seconds using Open Broadcaster Software (OBS)~\footnote{\url{https://obsproject.com}}.
Each glitchy clip follows a consistent temporal protocol: it begins with normal behavior; roughly one-third into the video, the temporal glitch is triggered and remains active for several seconds; and normal behavior resumes shortly before the end.
This protocol ensures that each glitchy video contains both normal and anomalous temporal segments, which is important for evaluating whether a model can detect temporally localized failures rather than merely memorize scene content.
It also better reflects realistic gameplay footage, where glitches typically occur during only part of a clip rather than throughout the entire video.
In total, \tempglitch contains 750 glitchy videos (150 per category) and 750 glitch-free videos, with 150 clean clips paired to each category. 
Section~\ref{sec:more_samples} shows more data examples per category of \tempglitch.

\section{Experiments}
\label{sec:experiments}
\paragraph{VLMs.} 
We evaluate a total of 12 VLMs on \tempglitch, including 6 proprietary models and 6 open-weight models.
The proprietary models are drawn from three leading model families: GPT~\cite{singh2025openai}, Gemini~\cite{team2023gemini}, and Claude~\cite{anthropic2024claude}.
The open-weight models cover the Qwen~\cite{bai2023qwen}, Gemma~\cite{gemmateam2025gemma3technicalreport}, and Ministral~\cite{liu2026ministral} families.
For proprietary models, we use the official APIs.
For open-weight VLMs, we use the public Unsloth framework\footnote{\url{https://unsloth.ai}}. 
See Section~\ref{sec:VLM_detail} for VLM details.
All experiments with open-weight models are run on 4 NVIDIA L40S GPUs.

\paragraph{Video prompting protocol.}
All models process videos as sequences of sampled frames except the Gemini family as Gemini models are the only ones to support direct video input.
To study the effect of temporal sampling, we extract frames using FFmpeg~\cite{ffmpeg} under multiple strategies, including 1 frame per second (1 FPS) and 5 frames per second (5 FPS).
Due to VLM input limits, we cap each video at a maximum of 50 extracted frames.
For each video, we ask the VLM to predict whether the video is glitchy or glitch-free based only on the provided visual evidence.
The model also returns a confidence score and a short reasoning explanation for its decision.
We require all responses to follow a valid JSON schema.
If a response is invalid or contains malformed fields, we retry the query until a valid JSON output is obtained.
Detailed prompt templates are provided in Section~\ref{sec:prompt_detail}.

\paragraph{Evaluation metrics.}
We evaluate glitch detection as a binary classification task, where each input is labeled as glitchy or glitch-free.
We report standard classification metrics: accuracy, F1, precision, and recall.
We use accuracy and F1 as the primary headline metrics for comparing models.
Precision and recall are reported as diagnostic metrics rather than standalone optimization targets, helping us interpret whether performance differences are driven by false positives or false negatives.

\section{Results}
\begin{table*}[t]
\centering
\caption{Performance of various VLMs on \tempglitch. All values are percentages. A dash indicates that the corresponding input type was not evaluated for that model.}
\label{tab:temporal_glitch_results}
\setlength{\tabcolsep}{3.5pt}
\begin{tabular}{lrrrrrrrrrrrr}
\toprule
& \multicolumn{4}{c}{Video} 
& \multicolumn{4}{c}{1 FPS} 
& \multicolumn{4}{c}{5 FPS} \\
\cmidrule(lr){2-5}
\cmidrule(lr){6-9}
\cmidrule(lr){10-13}
Model 
& Acc. & Prec. & Rec. & F1
& Acc. & Prec. & Rec. & F1
& Acc. & Prec. & Rec. & F1 \\
\midrule
Gemini 3.1 Pro 
& 49.7 & 49.8 & 57.6 & 53.4
& -- & -- & -- & --
& -- & -- & -- & -- \\

Gemini 3 Flash 
& 50.3 & 50.2 & 87.2 & 63.7
& -- & -- & -- & --
& -- & -- & -- & -- \\

\midrule
Claude Opus
& -- & -- & -- & --
& 51.0 & 53.0 & 17.5 & 26.3
& 50.2 & 50.6 & 16.7 & 25.1 \\

Claude Haiku
& -- & -- & -- & --
& 49.7 & 43.9 & 2.4 & 4.6
& 51.3 & 67.2 & 5.2 & 9.7 \\

GPT 5.5
& -- & -- & -- & --
& 50.5 & 58.0 & 3.9 & 7.2
& 51.1 & 68.1 & 4.3 & 8.0 \\

GPT 5.4 Mini
& -- & -- & -- & --
& \textbf{52.4} & 57.4 & 18.5 & 28.0
& \textbf{54.1} & 60.7 & 23.1 & 33.4 \\

\midrule
Qwen3.6 27B
& -- & -- & -- & --
& 50.5 & 50.5 & 54.3 & 52.3
& 50.5 & 50.4 & 54.5 & 52.4 \\

Qwen3VL 8B
& -- & -- & -- & --
& 50.7 & \textbf{69.2} & 2.4 & 4.6
& 50.1 & \textbf{100.0} & 0.3 & 0.5 \\

Gemma4 31B
& -- & -- & -- & --
& 52.3 & 53.1 & 39.7 & 45.5
& 53.0 & 53.7 & 43.5 & 48.0 \\

Gemma4 4B
& -- & -- & -- & --
& 50.0 & 0.0 & 0.0 & 0.0
& 50.0 & 0.0 & 0.0 & 0.0 \\

Ministral3 14B
& -- & -- & -- & --
& 51.9 & 68.8 & 7.1 & 12.8
& 52.6 & 67.9 & 9.9 & 17.2 \\

Ministral3 3B
& -- & -- & -- & --
& 50.2 & 50.1 & \textbf{96.9} & \textbf{66.1}
& 53.0 & 52.8 & \textbf{56.9} & \textbf{54.8} \\

\bottomrule
\end{tabular}
\end{table*}
Table~\ref{tab:temporal_glitch_results} summarizes results across 12 VLMs on \tempglitch. We highlight key findings and examine VLM
strengths and limitations in the remainder of this section.

\subsection{Current VLMs achieve near chance performance on \tempglitch}

Table~\ref{tab:temporal_glitch_results} shows that current VLMs struggle substantially on \tempglitch.
Because the benchmark is balanced between glitchy and glitch-free videos, an accuracy near 50\% indicates near-random discrimination.
Across all evaluated settings, the best accuracy is only 54.1\%, achieved by GPT 5.4 Mini under 5 FPS input.
Most models remain between 49\% and 53\% accuracy, regardless of whether they are proprietary or open-weight models.
This indicates that temporal glitch detection remains challenging even for strong recent VLMs.

The low accuracy is especially notable because the task is binary: models only need to decide whether a gameplay video is glitchy or glitch-free.
Unlike spatial glitches, which may be visible in a single frame, temporal glitches require reasoning about how the scene evolves over time.
The near-chance performance therefore suggests that current VLMs still lack robust temporal grounding.

\subsection{High recall often comes from over-predicting glitches}

Several models achieve high recall, but this often reflects a strong bias toward predicting videos as glitchy rather than reliable temporal understanding.
For example, Gemini 3 Flash with raw-video input achieves 87.2\% recall and 63.7\% F1, but its precision is only 50.2\% and its accuracy is only 50.3\%.
Similarly, Ministral3 3B under 1 FPS input reaches the highest recall of 96.9\% and the highest F1 of 66.1\%, but its precision is only 50.1\% and its accuracy is 50.2\%.
These results indicate that high F1 can be misleading on \tempglitch.
A model can obtain high recall by labeling nearly all videos as glitchy, but such behavior is not useful for practical QA because it would generate many false alarms.
Thus, \tempglitch exposes not only whether models can detect temporal glitches, but also whether they can do so without confusing normal gameplay variation with glitchy behavior.

\subsection{Many models miss almost all temporal glitches}

Other models fail in the opposite direction by predicting most videos as glitch-free.
For example, GPT 5.5 achieves relatively high precision under both 1 FPS and 5 FPS input (58.0\% and 68.1\%), but its recall remains extremely low (3.9\% and 4.3\%).
Claude Haiku also has very low recall under both 1 FPS and 5 FPS (2.4\% and 5.2\%).
Qwen3VL 8B becomes even more conservative under 5 FPS, reaching 100.0\% precision but only 0.3\% recall and 0.5\% F1.
Gemma4 4B collapses completely to the clean class, with 0.0\% recall and 0.0\% F1 under both frame-sampling settings.

Together with the high-recall models, these results reveal two dominant failure modes: some VLMs over-predict glitches, while others miss nearly all temporal glitches (despite being relatively good at game glitch detection that has a mix of spatial and temporal glitches~\cite{taesirivideogameqa}).

\subsection{Denser frame sampling provides limited and inconsistent gains}
Increasing the sampling rate from 1 FPS to 5 FPS can make short-lived glitches more visible in the sampled frames, see  Figure~\ref{fig:1vs5fps}.
However, this increased visibility does not translate into reliable performance gains across models.
GPT 5.4 Mini improves from 52.4\% to 54.1\% accuracy and from 28.0\% to 33.4\% F1.
Gemma4 31B improves from 52.3\% to 53.0\% accuracy and from 45.5\% to 48.0\% F1.
Ministral3 14B also improves from 12.8\% to 17.2\% F1.

However, denser sampling does not reliably improve all models.
Claude Opus slightly decreases from 26.3\% to 25.1\% F1, and Qwen3VL 8B drops sharply from 4.6\% to 0.5\% F1.
Qwen3.6 27B remains almost unchanged, with F1 changing only from 52.3\% to 52.4\%.
These results suggest that simply providing more frames is insufficient, even when denser sampling captures the glitchy moment. 
Temporal glitch detection requires models to integrate ordered visual evidence, not merely observe more isolated frames.

\subsection{Even raw-video input does not solve temporal glitch detection}

We additionally evaluate direct raw-video input (which is supported only by Gemini models).
The results show that raw-video input does not eliminate the difficulty of temporal glitch detection.
Gemini 3.1 Pro obtains 49.7\% accuracy and 53.4\% F1, while Gemini 3 Flash obtains 50.3\% accuracy and 63.7\% F1.
Although Gemini 3 Flash achieves high recall (87.2\%), its precision remains near chance (50.2\%), indicating strong over-prediction of the glitchy class.

This result is important because it separates input format from temporal reasoning capability.
Even when a model can directly process video input, it may still fail to distinguish true temporal glitches from normal gameplay dynamics.
Thus, accepting raw video as input does not necessarily imply robust video-level anomaly reasoning.

\subsection{Model scale alone does not explain performance}

The results do not show a simple relationship between model size and temporal glitch detection performance.
Within some families, larger models perform better.
For example, Gemma4 31B substantially outperforms Gemma4 4B, which predicts no glitches and obtains 0.0\% F1.
Qwen3.6 27B is also much more balanced than Qwen3VL 8B, achieving approximately 52\% F1 under both 1 FPS and 5 FPS settings.

However, larger models are not consistently better.
GPT 5.4 Mini outperforms GPT 5.5 under both 1 FPS and 5 FPS.
Ministral3 3B obtains much higher F1 than Ministral3 14B, although largely because it over-predicts glitches.
These patterns suggest that temporal glitch detection depends strongly on model family, prompting behavior, and calibration, rather than parameter scale alone.

\subsection{\tempglitch exposes distinct VLM failure modes}

Overall, \tempglitch reveals several distinct weaknesses in current VLMs.
Some models are overly conservative and miss most temporal glitches, as seen in GPT 5.5, Qwen3VL 8B, and Gemma4 4B.
Other models are overly sensitive and classify many clean videos as glitchy, as seen in Gemini 3 Flash and Ministral3 3B.
A third group, including Qwen3.6 27B and Gemma4 31B, achieves more balanced behavior but still remains close to chance accuracy.

These results show that \tempglitch is not merely a difficult leaderboard benchmark.
It is diagnostic: models with similar accuracy can fail in very different ways.
The benchmark therefore provides a useful testbed for studying temporal reasoning and robust gameplay understanding in VLMs.

In summary, the results on \tempglitch show that temporal glitch detection remains an open challenge for current VLMs.
Most models remain near chance accuracy, and higher F1 scores often reflect prediction bias rather than reliable detection.
Denser frame sampling and raw-video input provide limited gains, and model scale alone does not solve the task.
Together, these findings demonstrate the value of \tempglitch as a benchmark for temporally grounded gameplay glitch detection.

\section{Implications for Practical QA}
These failure modes have direct implications for how VLMs should be integrated into practical QA workflows.
A model that over-predicts glitches may appear useful under recall-oriented metrics, but it would require human testers to inspect a large number of false alarms.
Conversely, a model that is highly conservative may reduce review burden but would fail to surface many temporal glitches that are costly to reproduce later.
More importantly, glitch detection systems should not treat all glitches as a uniform detection problem.
Instead, they should process spatial and temporal glitches separately, using detection mechanisms that are specialized for the type of glitch being evaluated.

Our results also indicate that simply feeding sampled frames or raw videos into a VLM is insufficient for reliable temporal glitch detection. Current VLMs appear to lack task-specific knowledge about temporal gameplay glitches and often fail to identify whether a sequence of frames reflects normal game dynamics or an actual temporal error.
Future systems may therefore need to fine-tune VLMs on temporal glitch data or integrate explicit temporal reasoning components with VLMs.
Such components could help models track object states across frames and provide explanations grounded in the temporal evidence that supports a predicted glitch.
\tempglitch supports this type of analysis because each glitch category is paired with visually similar clean gameplay, making it difficult for a model to succeed by relying only on superficial scene cues.

\section{Limitations}
\label{sec:limitations}
The current dataset is built using the Godot engine and two base game environments. Although this design enables controlled evaluation and balanced coverage across temporal glitch types, it does not capture the full diversity of commercial games, rendering pipelines, camera systems, and gameplay mechanics. Future versions of the benchmark should extend to more games, engines, assets, and visual styles to evaluate whether model behavior generalizes beyond the current controlled setting.

\section{Conclusion}
In this paper, we introduced \tempglitch, the first benchmark for evaluating temporal glitch detection in gameplay videos. The design of \tempglitch is motivated by the distinction between spatial glitches, which can often be detected from a single frame, and temporal glitches, which require reasoning over ordered visual evidence. 
\tempglitch covers five temporal glitch types and pairs glitchy videos with matched glitch-free videos, enabling reliable binary evaluation and category-level analysis.
Our evaluation of 12 proprietary and open-weight VLMs shows that temporal glitch detection remains a major challenge. All studied models perform close to chance accuracy, and their failures often reflect strong prediction bias: some models miss most glitches, while others over-predict glitches on clean gameplay. Denser frame sampling, raw-video input, and model scale do not help improve performance. These findings highlight the need for dedicated methods that explicitly target temporal reasoning, calibration, and robust video-level understanding. In summary, \tempglitch fills an important gap in the evaluation of VLMs for video game QA by providing a dedicated benchmark for temporally grounded visual glitch detection.

{\small
\bibliographystyle{plain}
\bibliography{references.bib}
}






\newpage 

\appendix

\section{VLM Inference Details}
\label{sec:VLM_detail}
This section describes the inference providers and model-specific inference settings used in our benchmark.

\begin{table}[h]
\centering
\caption{Inference model providers and versions.}
\label{tab:inference}
\begin{tabular}{lll}
\toprule
\textbf{Model Name} & \textbf{Provider} & \textbf{Version} \\
\midrule
 Claude Opus 4.7 &\href{https://platform.claude.com/}{Claude}  &claude-opus-4-7  \\
 Claude Haiku 4.5 &\href{https://platform.claude.com/}{Claude}  &claude-haiku-4-5-20251001  \\
 GPT 5.5 &\href{https://openai.com/}{OpenAI}  &gpt-5.5-2026-04-23  \\
 GPT-5.4 Mini &\href{https://openai.com/}{OpenAI}  &gpt-5.4-mini-2026-03-17  \\
 Gemini 3.1 Pro &\href{https://ai.google.dev/gemini-api}{Gemini}  & gemini-3.1-pro-preview-02-26 \\
 Gemini 3 Flash &\href{https://ai.google.dev/gemini-api}{Gemini}  &gemini-3-flash-preview-12-25  \\
 Qwen3.6 27B   &\href{https://unsloth.ai/}{Unsloth}  &Qwen3.6-27B \\
 Qwen3VL 8B &\href{https://unsloth.ai/}{Unsloth}  &Qwen3-VL-8B-Instruct-unsloth-bnb-4bit  \\
 Gemma4 31B &\href{https://unsloth.ai/}{Unsloth}  &gemma-4-31B-it  \\
 Gemma4 E4B &\href{https://unsloth.ai/}{Unsloth}  &gemma-4-E4B-it  \\
 Ministral3 14B &\href{https://unsloth.ai/}{Unsloth}  &Ministral-3-14B-Instruct-2512  \\
 Ministral3 3B &\href{https://unsloth.ai/}{Unsloth}  &Ministral-3-3B-Instruct-2512  \\
\bottomrule
\end{tabular}
\end{table}

\section{VLM Prompting Details}
\label{sec:prompt_detail}
This section summarizes the prompts used for constructing the spatial--temporal glitch split in Section~\ref{sec:intro} and for evaluating VLMs on \tempglitch.

To separate existing samples from the PhysGame dataset into spatial and temporal glitches, we use the prompt shown in Figure~\ref{fig:prompt_1}. The prompt asks the VLM to produce a glitch/non-glitchy prediction, assign a glitch type, provide a confidence score, and give a short explanation grounded in the visual evidence. Using these responses, we first select candidate spatial and temporal glitch samples with high confidence scores. We then manually inspect the selected candidates and remove false positives, yielding the final 50-sample subsets for both spatial and temporal glitches. These subsets are used to evaluate which type of glitch is easier for VLMs to detect.

For evaluation on our \tempglitch dataset, we use the binary prompt shown in Figure~\ref{fig:prompt_binary}. This prompt asks the model to determine whether a gameplay video contains a glitch, allowing us to assess the ability of VLMs to detect temporal glitches. 

\begin{figure}[t]
\centering
\begin{tcolorbox}[
    width=\linewidth,
    colback=gray!2,
    colframe=gray!45,
    boxrule=0.7pt,
    arc=3pt,
    left=6pt,
    right=6pt,
    top=6pt,
    bottom=6pt,
    title=\textbf{Prompt Template for Spatial--Temporal Glitch Classification},
    colbacktitle=gray!25,
    coltitle=black,
    fonttitle=\bfseries
]

\begin{tcolorbox}[
    colback=blue!4,
    colframe=blue!60!black,
    boxrule=0.6pt,
    arc=2pt,
    title=\texttt{SYSTEM\_PROMPT},
    fonttitle=\bfseries\ttfamily\small,
    coltitle=white,
    colbacktitle=blue!65!black
]
{\ttfamily\small
You are evaluating keyframes from one gameplay video.

Decide whether the video shows a real gameplay bug/glitch based only on the frames.

Definitions:

- spatial bug:  a bug that is mainly visible in one or more still frames, such as clipping, floating, geometry overlap, missing or distorted textures, broken shadows, or static abnormal lighting/rendering.

- temporal bug: a bug that cannot be reliably determined from a single frame and mainly depends on changes over time, such as flickering, blinking, repeated popping, abnormal acceleration across frames.
- none: no visible bug in the frames.
        
If the frames are not sufficient to support a confident bug-type decision, use "unsure".
Do not use any outside knowledge. Use only visible evidence from the provided frames.
}
\end{tcolorbox}

\vspace{0.4em}

\begin{tcolorbox}[
    colback=orange!5,
    colframe=orange!75!black,
    boxrule=0.6pt,
    arc=2pt,
    title=\texttt{USER\_PROMPT},
    fonttitle=\bfseries\ttfamily\small,
    coltitle=white,
    colbacktitle=orange!75!black
]
{\ttfamily\small
These images are keyframes from a single gameplay video, in chronological order.

Return a JSON object with:

- is\_buggy: boolean

- bug\_type: one of ["spatial", "temporal", "none", "unsure"]

- confidence: number from 0 to 1

- explanation: short explanation grounded in the frames.
}
\end{tcolorbox}

\end{tcolorbox}
\caption{Prompt template used for constructing the spatial--temporal glitch subsets.}
\label{fig:prompt_1}
\end{figure}

\begin{figure}[!h]
\centering
\begin{tcolorbox}[
    width=\linewidth,
    colback=gray!2,
    colframe=gray!45,
    boxrule=0.7pt,
    arc=3pt,
    left=6pt,
    right=6pt,
    top=6pt,
    bottom=6pt,
    title=\textbf{Prompt Template for Glitch Detection},
    colbacktitle=gray!25,
    coltitle=black,
    fonttitle=\bfseries
]

\begin{tcolorbox}[
    colback=blue!4,
    colframe=blue!60!black,
    boxrule=0.6pt,
    arc=2pt,
    title=\texttt{SYSTEM\_PROMPT},
    fonttitle=\bfseries\ttfamily\small,
    coltitle=white,
    colbacktitle=blue!65!black
]
{\ttfamily\small
You are evaluating keyframes from one gameplay video. Decide whether the video shows a real gameplay bug/glitch based only on the frames. Do not use any outside knowledge. Use only visible evidence from the provided frames.
}
\end{tcolorbox}

\vspace{0.4em}

\begin{tcolorbox}[
    colback=orange!5,
    colframe=orange!75!black,
    boxrule=0.6pt,
    arc=2pt,
    title=\texttt{USER\_PROMPT},
    fonttitle=\bfseries\ttfamily\small,
    coltitle=white,
    colbacktitle=orange!75!black
]
{\ttfamily\small
These images are keyframes from a single gameplay video, in chronological order.

Return a JSON object with:

- is\_buggy: boolean

- confidence: number from 0 to 1

- explanation: short explanation grounded in the frames
}
\end{tcolorbox}

\end{tcolorbox}
\caption{Prompt template for video-based glitch detection.}
\label{fig:prompt_binary}
\end{figure}

\section{Samples for Different Temporal Glitch Types}
\label{sec:more_samples}
\begin{figure}[t]
  \centering
  \includegraphics[width=\textwidth]{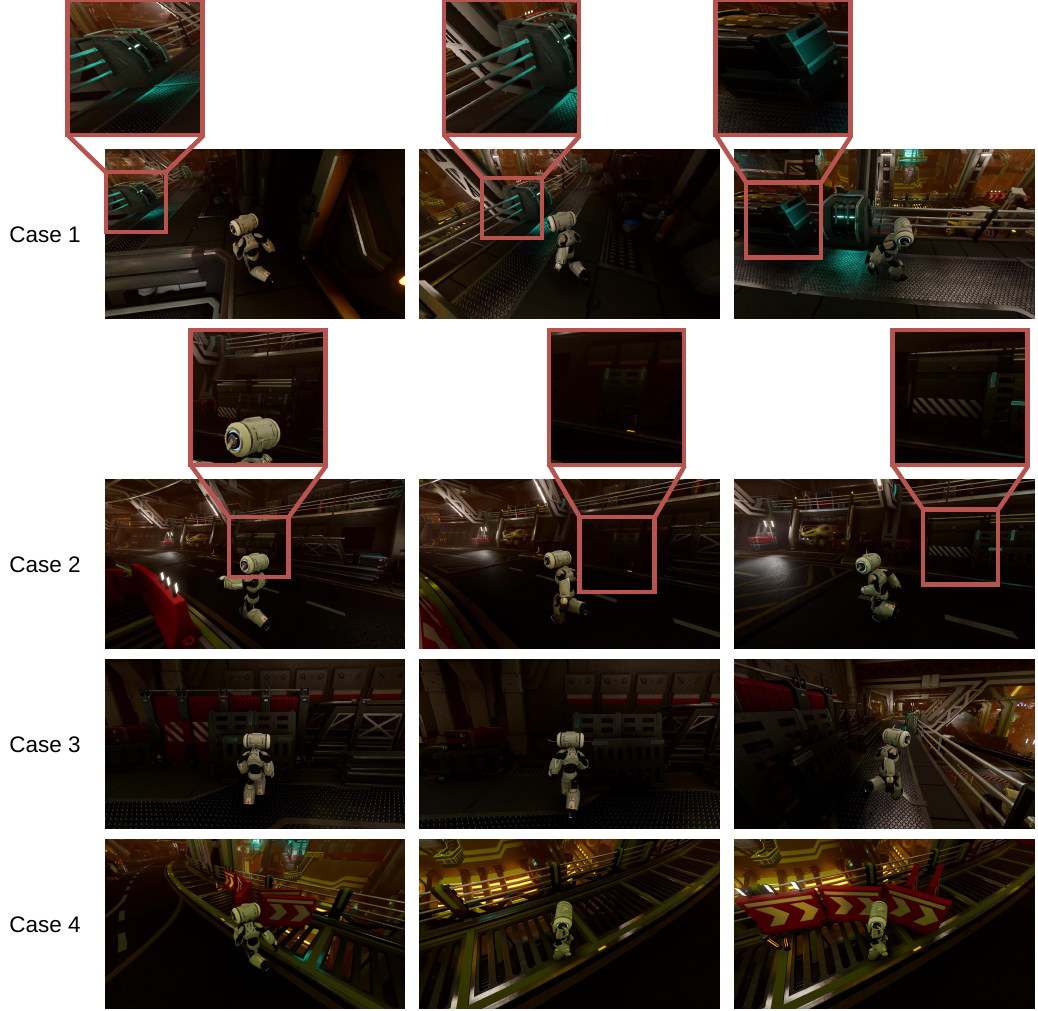}
  \caption{Examples of Blinking glitches.}
  \label{fig:data_blinking}
\end{figure}

\begin{figure}[h]
  \centering
  \includegraphics[width=\textwidth]{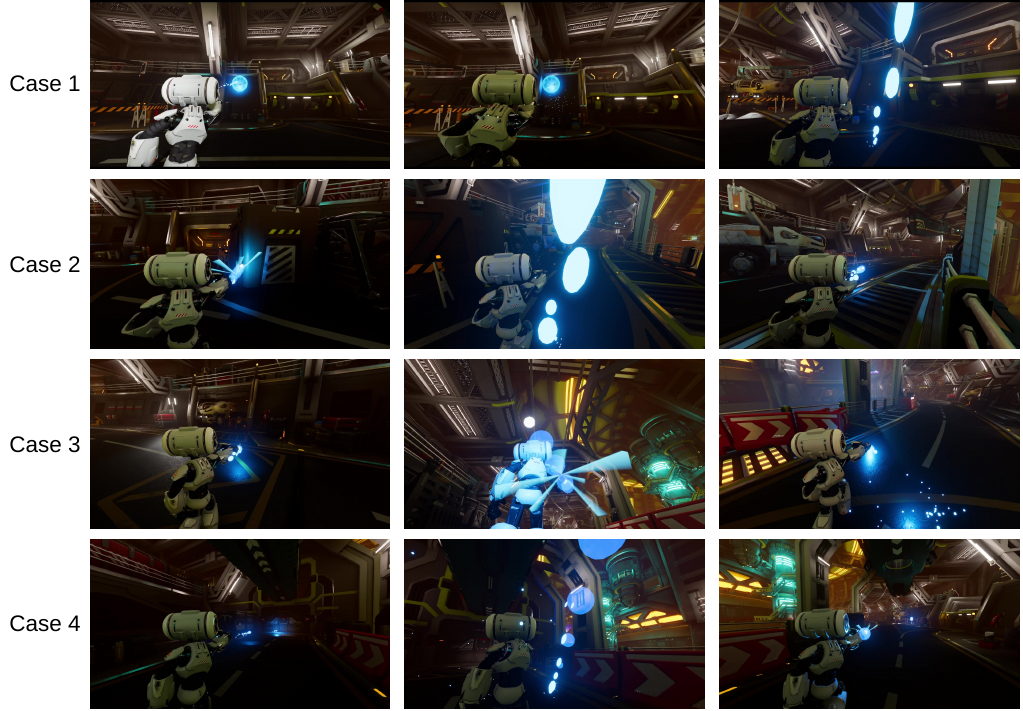}
  \caption{Examples of Shooting glitches.}
  \label{fig:data_shooting}
\end{figure}

\begin{figure}[h]
  \centering
  \includegraphics[width=\textwidth]{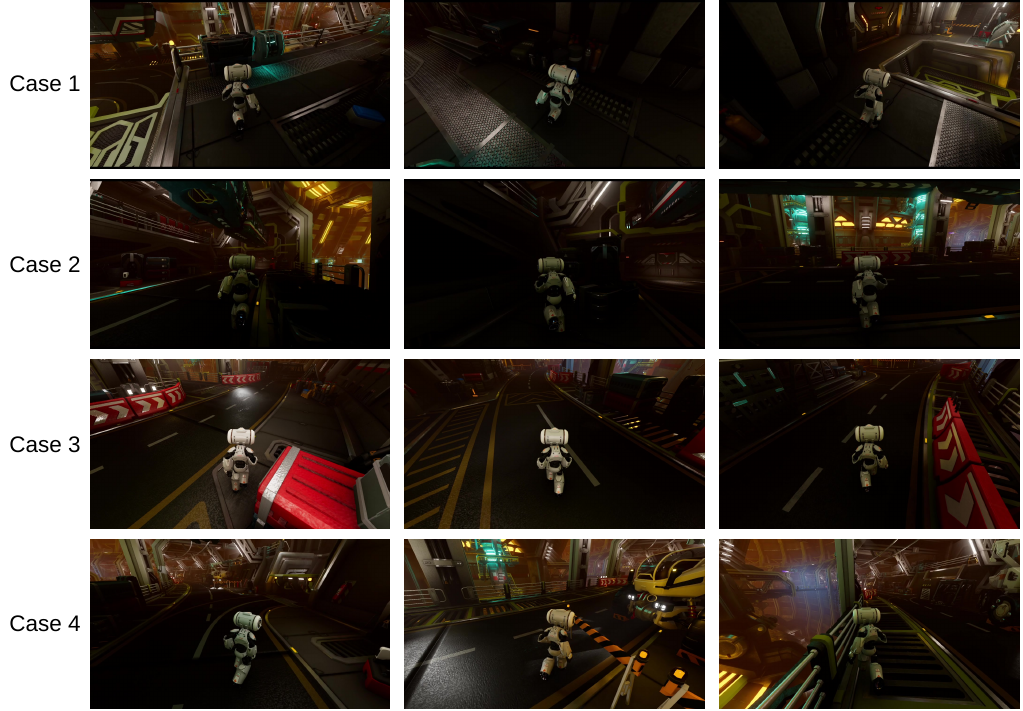}
  \caption{Examples of velocity glitches.}
  \label{fig:data_velocity}
\end{figure}

\begin{figure}[h]
  \centering
  \includegraphics[width=\textwidth]{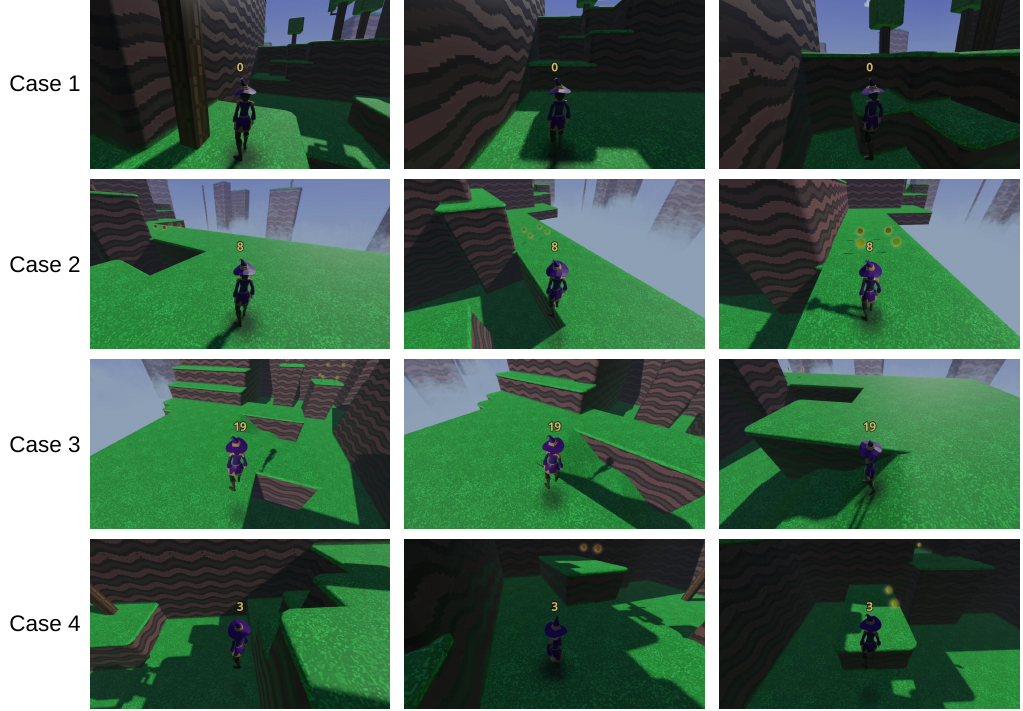}
  \caption{Examples of frozen glitches.}
  \label{fig:data_frozen}
\end{figure}

\begin{figure}[t]
  \centering
  \includegraphics[width=\textwidth]{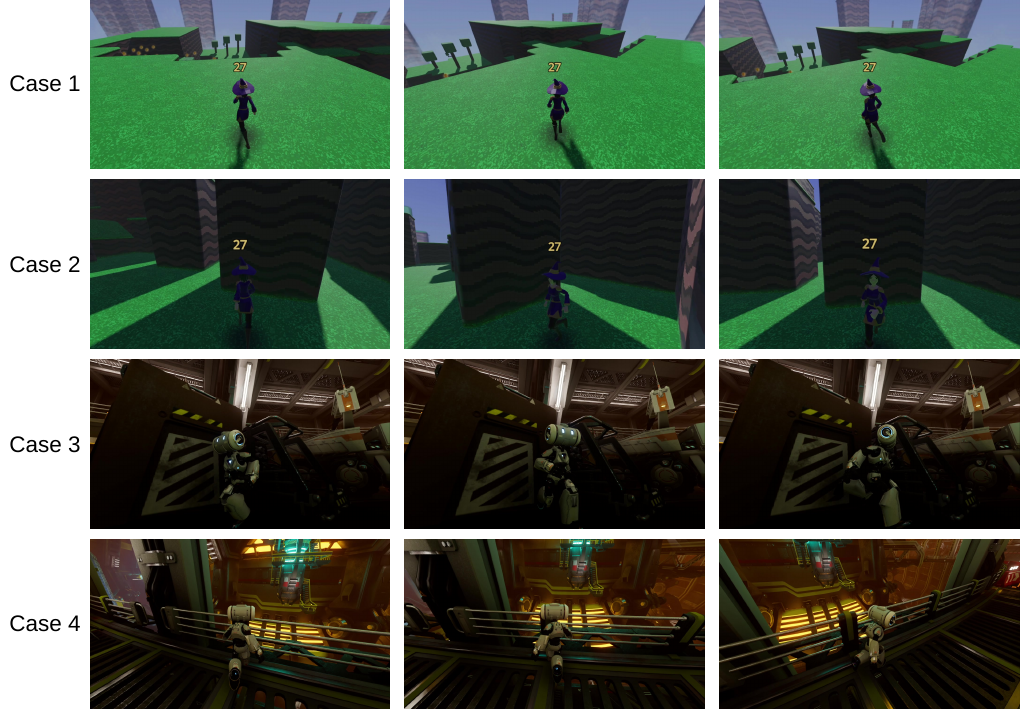}
  \caption{Examples of stuck-in-place glitches.}
  \label{fig:data_stuck}
\end{figure}

Figures~\ref{fig:data_blinking}--\ref{fig:data_stuck} show representative samples from the five temporal glitch categories. 
Figure~\ref{fig:data_blinking} shows blinking glitches, where an object or character intermittently appears and disappears across consecutive frames. 
Figure~\ref{fig:data_shooting} shows shooting-error glitches, where the firing effect or projectile trajectory is temporally misaligned with the weapon or expected target direction. 
Figure~\ref{fig:data_velocity} shows velocity glitches, where the character exhibits abnormal motion dynamics, such as abrupt displacement.
Figure~\ref{fig:data_frozen} shows frozen-animation glitches, where the character continues to move through the scene while its pose remains unnaturally fixed. 
Figure~\ref{fig:data_stuck} shows stuck-in-place glitches, where the character repeatedly attempts to move but remains constrained to nearly the same location. 
Together, these examples illustrate temporal failures that cannot be reliably diagnosed from a single frame and instead require reasoning over frame-to-frame changes.

\section{Effect of Frame Sampling Rate}
\begin{figure}[t]
  \centering
  \includegraphics[width=\textwidth]{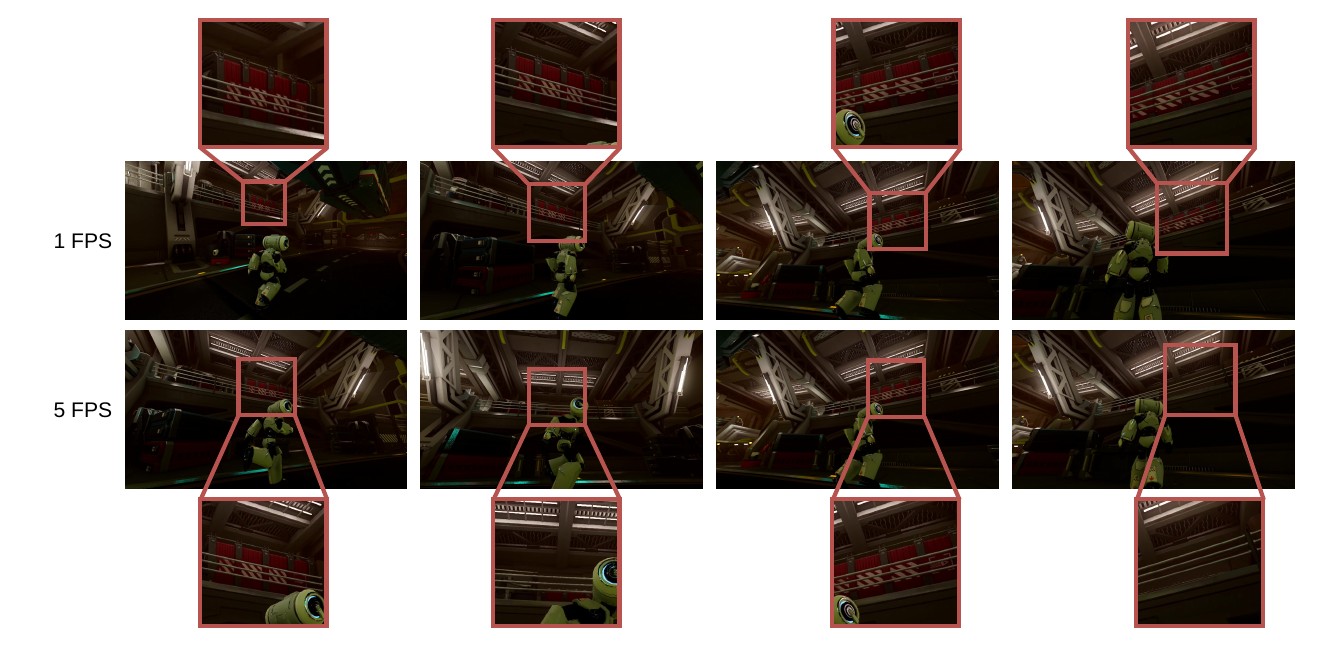}
  \caption{Frame sampling rate affects the visibility of blinking. A brief blinking glitch occurs as the character runs and turns its head. 
  The red boxes highlight the affected object. 
  Sampling at 1 FPS misses the glitch, whereas 5 FPS captures intermediate frames in which the blinking becomes visible.}
  \label{fig:1vs5fps}
\end{figure}
Frame sampling rate directly affects whether short-lived temporal glitches are visible in the sampled frames. 
Figure~\ref{fig:1vs5fps} illustrates this effect with a blinking glitch that occurs as the character runs and turns its head. 
The object highlighted by the red box briefly flickers, but this moment is missed when the video is sampled at 1 FPS. 
In contrast, sampling at 5 FPS captures intermediate frames in which the glitch is visible. 
However, despite exposing the glitch, denser frame sampling does not substantially improve VLM-based bug detection, with performance remaining close to chance. 
This suggests that simply increasing the number of sampled frames, even when the glitch itself is included, is insufficient for reliable temporal glitch understanding.

\section{Additional Performance Metrics}

\begin{table*}[t]
\centering
\caption{Confusion-matrix counts of various VLMs on \tempglitch. A dash indicates that the corresponding input type was not evaluated for that model.}
\label{tab:temporal_glitch_confusion}
\setlength{\tabcolsep}{3.5pt}
\begin{tabular}{lcccccccccccc}
\toprule
& \multicolumn{4}{c}{Video} 
& \multicolumn{4}{c}{1 FPS} 
& \multicolumn{4}{c}{5 FPS} \\
\cmidrule(lr){2-5}
\cmidrule(lr){6-9}
\cmidrule(lr){10-13}
Model 
& TP & TN & FP & FN
& TP & TN & FP & FN
& TP & TN & FP & FN \\
\midrule
Gemini 3.1 Pro 
& 432 & 314 & 436 & 318
& -- & -- & -- & --
& -- & -- & -- & -- \\

Gemini 3 Flash 
& 654 & 100 & 650 & 96
& -- & -- & -- & --
& -- & -- & -- & -- \\

\midrule
Claude Opus
& -- & -- & -- & --
& 131 & 634 & 116 & 619
& 125 & 628 & 122 & 625 \\

Claude Haiku
& -- & -- & -- & --
& 18 & 727 & 23 & 732
& 39 & 731 & 19 & 711 \\

GPT 5.5
& -- & -- & -- & --
& 29 & 729 & 21 & 721
& 32 & 735 & 15 & 718 \\

GPT 5.4 Mini
& -- & -- & -- & --
& 139 & 647 & 103 & 611
& 173 & 638 & 112 & 577 \\

\midrule
Qwen3.6 27B
& -- & -- & -- & --
& 407 & 351 & 399 & 343
& 409 & 348 & 402 & 341 \\

Qwen3VL 8B
& -- & -- & -- & --
& 18 & 742 & 8 & 732
& 2 & 750 & 0 & 748 \\

Gemma4 31B
& -- & -- & -- & --
& 298 & 487 & 263 & 452
& 326 & 469 & 281 & 424 \\

Gemma4 4B
& -- & -- & -- & --
& 0 & 750 & 0 & 750
& 0 & 750 & 0 & 750 \\

Ministral3 14B
& -- & -- & -- & --
& 53 & 726 & 24 & 697
& 74 & 715 & 35 & 676 \\

Ministral3 3B
& -- & -- & -- & --
& 727 & 26 & 724 & 23
& 427 & 368 & 382 & 323 \\

\bottomrule
\end{tabular}
\end{table*}
Table~\ref{tab:temporal_glitch_confusion} reports the confusion-matrix counts for each VLM and input setting. 
These results complement the aggregate metrics in Table~\ref{tab:temporal_glitch_results} by showing whether performance differences are driven by true glitch detections, correct clean-video rejections, false alarms, or missed glitches. 
This is particularly important for \tempglitch, since a model can obtain high recall by over-predicting the glitchy class while still producing many false positives on glitch-free videos.







\end{document}